\pdfoutput=1

\documentclass[11pt]{article}

\usepackage[final]{acl}
\usepackage{algorithm}
\usepackage{algorithmic}
\usepackage{times}
\usepackage{booktabs}
\usepackage{multirow}
\usepackage{url}
\usepackage{latexsym}
\usepackage{tcolorbox}
\usepackage{colortbl}
\usepackage{color}
\usepackage{xcolor}
\usepackage{mdframed}
\definecolor{NavyBlue}{rgb}{0.0, 0.0, 0.5}
\definecolor{BlueGreen}{rgb}{0.0, 0.5, 0.5}
\definecolor{Plum}{rgb}{0.5, 0.0, 0.5}
\definecolor{ForestGreen}{rgb}{0.0, 0.26, 0.15}
\definecolor{Red}{rgb}{1.0, 0.0, 0.0}
\usepackage{enumitem}
\usepackage[T1]{fontenc}

\usepackage[utf8]{inputenc}

\usepackage{microtype}

\usepackage{inconsolata}
\usepackage{amsmath}
\usepackage{amssymb}
\usepackage{amsfonts}
\usepackage{geometry} 
\usepackage{graphicx}
\usepackage{epigraph}
\definecolor{ZSBaseline}{HTML}{ff8d13}
\definecolor{KDBaseline}{HTML}{bd00ff}
\definecolor{DABaseline}{HTML}{2782ed}
\definecolor{OurColor}{HTML}{36aa70}
\definecolor{UserExampleBg}{HTML}{ffffff}
\definecolor{UserExampleTitle}{HTML}{545f7f}
\newmdenv[
    roundcorner=5pt,
    backgroundcolor=UserExampleBg,
    linecolor=UserExampleTitle,
    outerlinewidth=0.5pt,
    frametitlebackgroundcolor=UserExampleTitle,
    frametitlefont={\bfseries\color{white}},
]{user_example}
\title{Reversal of Thought: Enhancing Large Language Models with Preference-Guided Reverse Reasoning Warm-up}

\author{
 \textbf{Jiahao Yuan\textsuperscript{1}\thanks{51275900024@stu.ecnu.edu.cn}},
 \textbf{Dehui Du\textsuperscript{1}\thanks{Corresponding Author: dhdu@sei.ecnu.edu.cn}},
 \textbf{Hao Zhang\textsuperscript{1}},
 \textbf{Zixiang Di\textsuperscript{1}},
 \textbf{Usman Naseem\textsuperscript{2}}
\\
 \textsuperscript{1}East China Normal University
 \\
 \textsuperscript{2}Macquarie University
}

\begin{document}
\maketitle
\begin{abstract}
Large language models (LLMs) have shown remarkable performance in reasoning tasks but face limitations in mathematical and complex logical reasoning. Existing methods to improve LLMs' logical capabilities either involve traceable or verifiable logical sequences that generate more reliable responses by constructing logical structures yet increase computational costs, or introduces rigid logic template rules, reducing flexibility. In this paper, we propose Reversal of Thought (RoT), a plug-and-play and cost-effective reasoning framework designed to enhance the logical reasoning abilities of LLMs during the warm-up phase prior to batch inference. RoT utilizes a \textit{Preference-Guided Reverse Reasoning} warm-up strategy, which integrates logical symbols for pseudocode planning through meta-cognitive mechanisms and pairwise preference self-evaluation to generate task-specific prompts solely through demonstrations, aligning with LLMs’ cognitive preferences shaped by RLHF. Through reverse reasoning, we utilize a \textit{Cognitive Preference Manager} to assess knowledge boundaries and further expand LLMs’ reasoning capabilities by aggregating solution logic for known tasks and stylistic templates for unknown tasks. Experiments across various tasks demonstrate that RoT surpasses existing baselines in both reasoning accuracy and efficiency. 
\end{abstract}

\section{Introduction}
\begin{figure}[ht]
  \includegraphics[width=\linewidth]{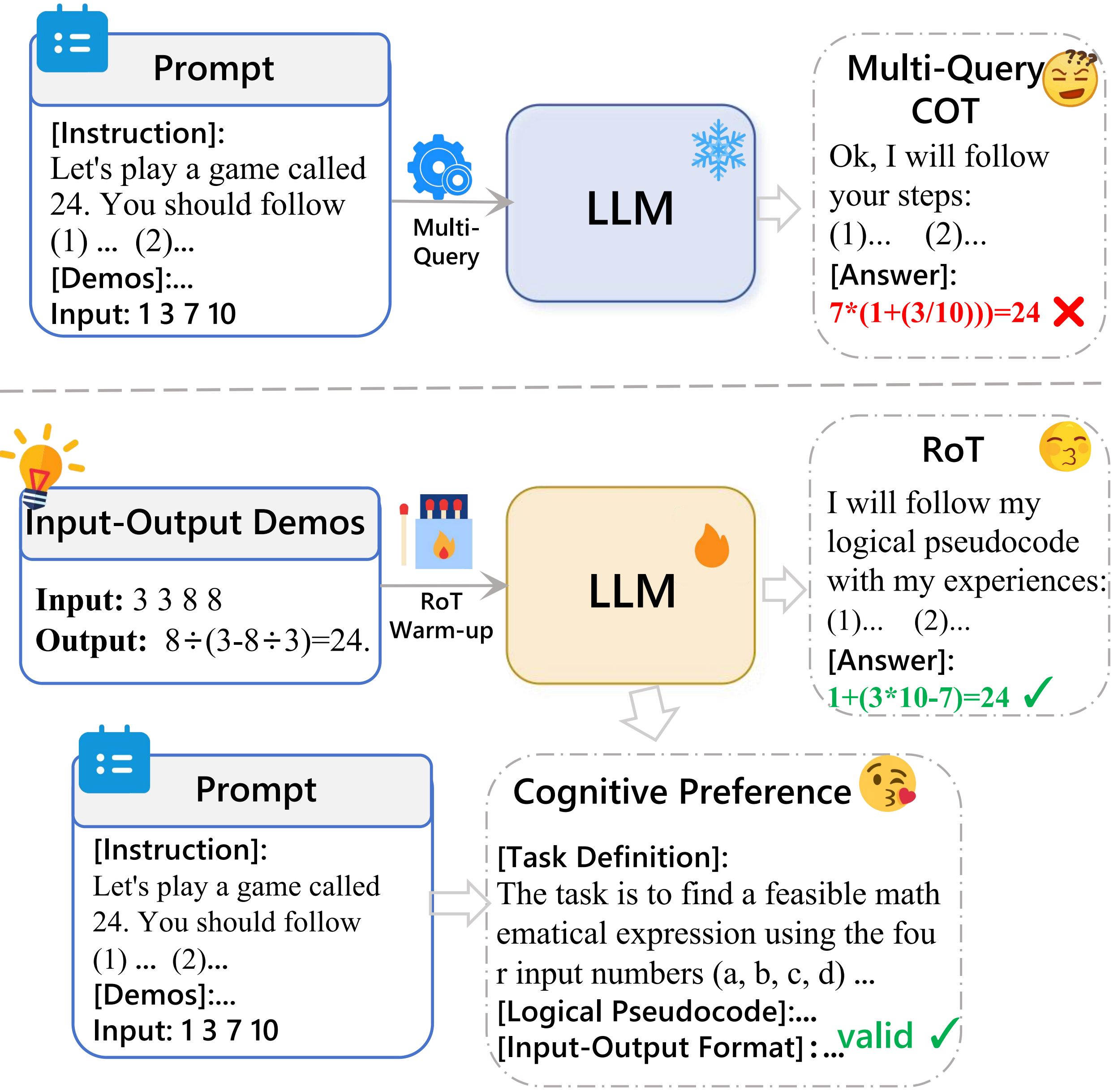}
  \caption{Comparison between CoT \cite{yao2024tree,besta2024graph,yang2024buffer} and Reversal of Thought (RoT)}
  \label{fig:intro}
\end{figure}

Large language models (LLMs) like Qwen \cite{bai2023qwen}, Llama \cite{dubey2024llama}, and GPT-4 \cite{achiam2023gpt} have demonstrated remarkable performance in various reasoning tasks via single-step prompting with few shots upon scaling model size \cite{plaat2024reasoning} but remain restricted in mathematical and intricate logical reasoning domains \cite{arkoudas2023gpt,stechly2023gpt}, which has spurred more effective multi-step Chain-of-Thought (CoT) prompting \cite{wei2022chain} approaches for activating step-by-step logical capabilities. However, LLMs are prone to unfaithfulness, resulting in cascaded intermediate errors \cite{bao2024llms,yang2024arithmetic}.  

Recent studies have advanced CoT to guide LLMs, mainly through either multi-step prompting such as introducing planning-and-solve \cite{plaat2024reasoning,yang2024buffer}, self-consistency \cite{narangself,wang2024make} and recursive reasoning process \cite{lee2023recursion,yu2024teaching} through Tree-of-Thought (ToT) \cite{yao2024tree}, Graph-of-Thought (GoT) \cite{besta2024graph}, or multi-role \cite{zhangcumulative,suzgun2024meta} to enhance logical capabilities and mitigate hallucination, yet this has stealthily increased inference cost due to the multi-step inference. Buffer-of-Thought (BoT) \cite{yang2024buffer} attempts to reduce thinking steps by leveraging Retrieval-Augmented Generation (RAG) to retrieve gold thought templates from the buffer. However, it sacrifices flexibility due to the initialization of pre-set manual thought templates. Therefore, achieving accurate reasoning in LLMs while minimizing resource consumption remains a significant challenge.

In summary, existing methods primarily rely on multi-query CoT which injects knowledge~\cite{suzgun2024meta,plaat2024reasoning} or data structure~\cite{yao2024tree,besta2024graph} to optimize decisions making, and encounter three significant limitations: \textbf{(1) limitation in logical reasoning:} Despite attempting different logic data structures \cite{yao2024tree,besta2024graph,yang2024buffer}, an effective initiative Chain-of-Thought paradigm that suits and improves logical reasoning remains elusive \cite{bao2024llms}; \textbf{(2) unfaithfulness and cascaded errors:} Single-step or multi-step methods are liable to cause LLMs to output hallucinations, leading to cascading logic errors \cite{bao2024llms};
\textbf{(3) Trade-off between enhanced logic capabilities and resource consumption:} Recent CoT advancements via multi-step or multi-role prompting increase costs and achieving a balance between logical flexibility, accuracy, and cost is of great significance for practical application.

To address above limitations, inspired by meta-cognition \cite{fleur2021metacognition} and and cognitive preference \cite{uddin2021cognitive, zhou2023biasing, margatina2023active}, we propose Reversal of Thought (RoT), a plug-and-play and cost-effective framework that enables LLMs to explore cognitive preference on logical pseudocode solely using reverse prompting with given demos without additional task-related affirmations, as depicted in Figure~\ref{fig:intro}.
Our key contributions are as follows:
\begin{itemize}
    \item To the best of our knowledge, we are the first to introduce a reversal reasoning for cognitive preference that enhances logical reasoning in LLMs by combining meta-cognitive with cognitive preference, resulting in a more modular and cost-efficient framework for complex tasks.
    \item We propose a \textit{Preference-Guided Reverse Reasoning} framework that enhances LLMs' task cognition by employing a reverse reasoning warm-up strategy and preference-based self-evaluation to improve logical reasoning based on LLMs’ cognitive preferences.
    \item We introduce a \textit{Cognitive Preference Manager} to evaluate knowledge boundaries, enabling the automatic adaptation of cognitive preference styles for unknown logic tasks and efficient aggregation of solution logic for known tasks.

\end{itemize}

\section{Related Work}
\subsection{Chain-of-Thought (CoT)  Prompting}
Chain-of-Thought (CoT) prompting \cite{wei2022chain} has been proven to be a promising approach that incorporates an intermediate logic chain to enhance LLMs' logic. Recent studies primarily aimed at improving logical accuracy by external validation mechanisms \cite{bi2025cotkinetics} like symbolic reasoning \cite{cai2023large,pan2023logic}, stepwise verification including self-consistency \cite{narangself,yu2024teaching,wang2024make}, self-refine \cite{madaan2024self}, self-reflection \cite{renze2024self} and more hierarchical information such as Least-to-Most \cite{zhouleast}, Cumulative-Reasoning \cite{zhangcumulative} and Multi-experts \cite{suzgun2024meta} strategies, but faced challenges related to cumulative errors \cite{bao2024llms} or poor flexibility \cite{yang2024buffer}. Additionally, numerous studies also proposed more standardized recursive or backtracking branch forms from the logical data structure, including Tree-of-Thought (ToT) \cite{yao2024tree}, Graph-of-Thought (GoT) \cite{besta2024graph} and Buffer-of-Thought (BoT) \cite{yang2024buffer}. However, an efficient logical reasoning method that strikes a balance among reasoning accuracy, flexibility, and cost has yet to be discovered. Our method is activated through meta cognition \cite{fleur2021metacognition} by introducing reverse reasoning to form effective LLMs-taste prompts within cognitive preference \cite{uddin2021cognitive} for plan-and-solve with logical pseudocode at least.

\subsection{Knowledge Boundary for Enhancing Large Language Models}
Integrating knowledge boundary within LLMs has emerged as a prospective strategy for enhancing their ability to avoid reasoning hallucinations of unknown knowledge through knowledge boundary constraints which requires additional algorithmic efforts \cite{yin2024benchmarking,chenunlocking}, external graph knowledge \cite{tian2024theory}, and training consumption \cite{sun2024determlr}. Additionally, they focus on avoiding responses to unknown or incorrect prompts rather than proposing bold and proactive solutions to expand knowledge boundary in a heuristics without training. 
\begin{figure*}[ht]
  \centering
  \includegraphics[width=1\linewidth]{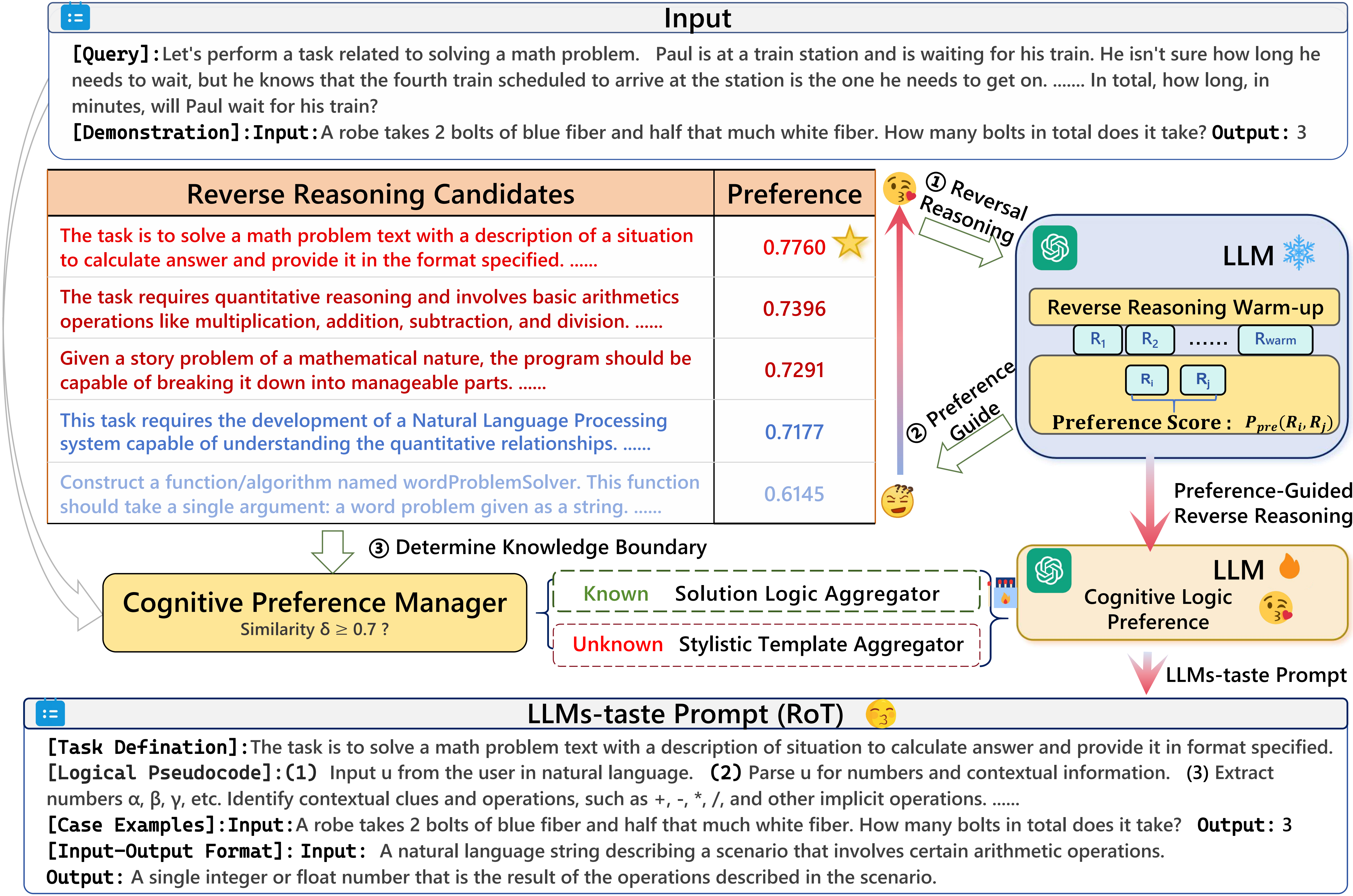}
  \caption{Architecture of Reversal-of-Thought (RoT). RoT comprises two primary components: \textit{Preference Guided Reverse Reasoning}, which enhances logical reasoning by activating LLMs' cognitive preferences, and \textit{Cognitive Preference Manager}, which assesses knowledge boundaries and adapts cognitive styles for various tasks. }
  \label{fig:model}
\end{figure*}
We proposed a prompt-based method utilizing LLMs pretrained knowledge boundary, inspired by meta cognition \cite{fleur2021metacognition} and cognitive preference for unknown knowledge \cite{uddin2021cognitive}. Our method conducts reverse prompting on probing knowledge through demonstrations to obtain LLMs-taste problem cognitions, aggregates and distills original prompt into cognitive preference version.

\section{Reversal of Thought}

\subsection{Overview}
\epigraph{\textit{Tell me and I forget. Teach me and I remember. Involve me and I learn.}}{\citet{franklin2005tell}}
\vspace{-0.6em}

As the aforementioned wisdom related to human cognitive learning implies, merely telling or teaching is inadequate \cite{bao2024llms}. Moreover, most LLMs have undergone extensive pre-training \cite{achiam2023gpt,bai2023qwen,dubey2024llama} and reinforcement learning from human feedback (RLHF) \cite{ouyang2022training}, instilling in LLMs a propensity for specific cognitive patterns, which manifests in two progressive layers of LLMs-taste description: \textbf{(1) Stylistic template:} encompassing grammatical and syntactic structures in descriptions for thinking problems. \textbf{(2) Solution logic:} comprising problem-solving reasoning and methodological cues. Therefore, \textit{Reversal of Thought (RoT)} involves answering the following two research questions (RQs):

\begin{itemize} 
\item \textit{\textbf{RQ1:} How to make LLMs output preference cognitive templates and logic for specific tasks and activate known cognitive boundaries?
\item \textbf{RQ2:} How to autonomously use cognitive templates with incorrect response to expand the possible knowledge boundaries?}
\end{itemize}

To activate and enhance LLMs logical flexibility, accuracy, and the ability to autonomously construct meta-cognition without training for logical reasoning, inspired by meta-cognition \cite{fleur2021metacognition} and cognitive preference \cite{uddin2021cognitive, zhou2023biasing, margatina2023active}, we introduce \textit{Reversal of Thought (RoT)}, a cost-effective paradigm that enables LLMs to first explore cognitive preference on logical pseudocode solely through given examples without additional task-oriented affirmation, activates the pre-trained known logic under \textit{Reverse Reasoning Warm-up} (detailed in section \ref{subsec:reverse}), and then optimizes the original prompt for LLMs-taste prompt via \textit{Cognitive Preference Manager} (detailed in section \ref{subsec:cognitive}) to determine the transfer of cognitive preference style for unknown logic template and aggregation of known solution logic, as depicted in Figure~\ref{fig:model}.
\begin{algorithm}[tb]\small
    \caption{Preference-Guided Reverse Reasoning (PGRR)}
    \label{alg:reverse}
    \begin{algorithmic}[1]
        \REQUIRE $P$: Initial prompt, $D$: Input-output demonstrations, $warm$: Number of warm iterations
        \STATE \textbf{(1) Reverse Reasoning Warm-up:}
        \FOR{$i = 1$ \textbf{to} $warm$}
            \STATE $R^{(i)} \gets \mathcal{M}_{\text{LLM}}(P_r, D, i)$ \COMMENT{Generate candidate responses}
            \STATE $P_{res}^i \gets \frac{1}{|R^{(i)}|} \sum_{R_{i,j} \in R^{(i)}} \exp(P(R_{i,j}| P_r, D))$
        \ENDFOR

        \STATE $R \gets \bigcup_{i=1}^{warm} R^{(i)}$ \COMMENT{Collect all responses}

        \textbf{(2) Pairwise Preference Evaluation:}
        \FOR{$i = 0$ \textbf{to} $warm-1$}
            \STATE $P_{pre}(R_{i+1} \succ R_i) \gets \exp(\mathcal{M}_{\text{LLM}}(P_{eval}, R_{i+1}, R_i))$
        \ENDFOR
        
        \FOR{$i = 0$ \textbf{to} $warm-1$}
            \FOR{$j = 0$ \textbf{to} $i-1$}
                \STATE $P_{pre}(R_i \succ R_j) \gets P_{pre}(R_{i} \succ R_{j})$ \COMMENT{Utilize preference transitivity}
            \ENDFOR
        \ENDFOR

        \textbf{(3) Preference-Guided Ranking:}
        \FOR{$i = 1$ \textbf{to} $warm$}
            \STATE $\bar{P}_{pre}(R_i) \gets \frac{1}{warm-1} \sum_{\substack{j=1 \\ j \neq i}}^{warm} P_{pre}(R_i \succ R_j)$
        \ENDFOR
        \STATE $P_{opt} \gets \arg \max_{R_i} \left( \frac{P_{res}^i + \bar{P}_{pre}(R_i)}{2} \right)$
        
        \RETURN $P_{opt}$ \COMMENT{Optimal LLMs-taste prompt}
    \end{algorithmic}
\label{reverse}
\end{algorithm}

\subsection{Reverse
Reasoning with Meta-cognition}
\label{subsec:reverse}
\paragraph{Preference-Guided Reverse Reasoning.}Inspired by RLHF \cite{ouyang2022training} utilizing preference data, and to derive high-cognitive preference prompt $P^*$ that enhance logical reasoning in LLMs, we propose a \textit{Preference-Guided Reverse Reasoning (PGRR)} framework (detailed in alogrithm~\ref{alg:reverse}) mapping input-output demonstrations $D$ from an initial prompt $P$ to an optimal LLM-taste prompt $P_{opt}$ from prompt candidates.

\subparagraph{(1) Reverse Reasoning Warm-up.} We query the LLM $\mathcal{M}_{\text{LLM}}$ with a reversal prompt and demonstrations $\{P_r, D\}$ (detailed in figure~\ref{prompt}) for $warm$ iterations, generating a set of prompt candidates for solution $R = \{R_1, R_2, \dots, R_{warm}\}$ and their corresponding average probabilities $P_{res}^i$:
\begin{center}
\begin{align}
R &= \bigcup_{i=1}^{\text{warm}} R^{(i)} = \bigcup_{i=1}^{\text{warm}} \mathcal{M_{LLM}}(P_r, D, i),
\end{align}

\begin{align}
P_{res}^i &= \frac{1}{|R_i|} \sum_{R_{i,j} \in R} \exp(P(R_{ij}| P_r, D)).
\end{align}
\end{center}

where $R^{(i)}$ represents the $i$-th generated response. $\mathcal{M}_{\text{LLM}}(P_r, D, i)$ is the model output based on the reversal prompt $P_r$ and demonstrations $D$ for the $i$-th iteration. $ P(R_{ij}|P_r,D) $ denotes log probability for each token $ R_{ij} \in R_i $ from LLMs.

\subparagraph{(2) Pairwise Preference Evaluation.}
To acquire the most LLMs-taste prompt, we pair candidate responses $R$ as data pairs $\{R_i, R_{i+1}\}$ where $i = 0, 1, \cdots, warm-1$ to calculate the relative preference $P(R_{i+1} \succ R_i)$ through LLM’s self-evaluation of its preference for $R_{i+1}$ over $R_i$, formally define as:
\begin{equation} 
\label{evaluate} 
P_{pre}(R_{i+1}\succ R_{i})=\exp(\mathcal{M_{LLM}}(P_{eval},R_{i+1},R_{i}))
\end{equation}
where $\mathcal{M_{LLM}}(P_{eval}, R_{i+1}, R_{i})$ represents that require LLM to select more preferred data through $P_{eval}$ with a structure as \textit{"Please choose your more preferred instruction (A/B): (A).$R_{i+1}$; (B).$R_{i}$".}

Following the principle of preference transitivity \cite{liu2024aligning}, we extend $P(R_{i+1} \succ R_i)$ to $P(R_i \succ R_j)$ to reduce computational cost from, thereby forming a preference matrix $P_{pre} \in \mathbb{R}^{warm \times warm}$, formally:
\begin{equation} 
\small
\label{pairwise} 
P_{pre}(R_i\succ R_j)=\begin{cases}
1&i=j\\\prod_{k=j}^{i-1}P_{pre}(R_{k + 1}\succ R_k)&i>j\\
1-{P_{pre}(R_j\succ R_i)}&i<j
\end{cases}
\end{equation}

\subparagraph{(3) Preference-Guided Ranking.} 
To identify the most LLMs-preferred and high-quality response, we compute each response $R_i$'s overall preference score $\bar{P}_{pre}(R_i)$, and averaging both average probabilities $P_{res}^i$ in matrix $P_{pre}$ and preference score $\bar{P}_{pre}(R_i)$ to obtain the best LLM-taste prompt $P_{opt}$:

\begin{small}
\begin{align}
\label{average_preference}
\bar{P}_{pre}(R_i) &= \frac{1}{warm-1} \sum_{\substack{j=1 \\ j \neq i}}^{warm} P_{pre}(R_i \succ R_j), \\
P_{opt} &= \arg \max_{R_i} \left( \frac{P_{res}^i + \bar{P}_{pre}(R_i)}{2} \right).
\end{align}
\end{small}

\paragraph{Reverse Logic for Meta-cognition.} 
Within reverse reasoning, we further follow meta-cognitive \cite{suzgun2024meta} using plan-and-solve by integrating logical algorithm pseudo-code to improve reasoning comprehension. And we incorporate fundamental mathematical logic symbols, including logical operators, quantifiers, inequalities and conditional statements, to facilitate model reasoning detailed in Appendix~\ref{subsec:prompt}.

\subsection{Cognitive Preference Manager}
\label{subsec:cognitive}
\begin{algorithm}[tb]\small
    \caption{Cognitive Preference Manager (CPM)}\label{alg:cpm}
    \begin{algorithmic}[1]
        \REQUIRE $P$: Original prompt, $P^*$: Reverse-reasoned prompt, $\mathcal{M}_{emb}$: Offline LLM embedding model, $\delta$: Similarity threshold
        \STATE $P_{task} \gets P , P^*_{task} \gets P^*$
        \STATE $s \gets \mathcal{M}_{emb}(P_{task}, P^*_{task})$ Calculate embedding similarity between $P_{\text{task}}$ and $P^*_{\text{task}}$.
        \IF{$s \geq \delta$}
            \STATE \textbf{Known detected:} Enhance and refine $P$ to optimized instructions $P_{final}$.
            \STATE Aggregate relevant task-specific knowledge.
            \RETURN $P_{final}$.
        \ELSE
            \STATE \textbf{UnKnown detected:} Adapt and expand $P_{task}$.
            \STATE Leverage cognitive preference templates $T$ and $P$ to generate optimized instructions $P_{final}$.
            \RETURN $P_{final}$.
        \ENDIF
    \end{algorithmic}
\end{algorithm}

\paragraph{Cognitive Preference Manager.} After reverse reasoning for LLMs-cognitive description $P^*$, We introduce an offline-deployed LLM embedding model $\mathcal{M}_{emb}$ to  assist \textit{Cognitive Preference Manager (CPM)} in determining whether reverse reasoning under reverse prompt $P_r$ and demonstrations $D$ reaches the knowledge boundary or cognitive error by calculating the similarity and setting a threshold $\delta$ (0.6$\sim$0.8 is recommended for optimal performance in 
distinguishing knowledge boundaries) between orginal task defination $P_{task}$ from $P$ and LLMs-cognitive task defination $P^*_{task}$ from $P^*$, and finally get a cognitive signal $S_{cog}$, formally:

\begin{small}
\begin{equation}
\label{eq:similarity}
S_{\text{cog}} = 
\begin{cases} 
unknown & , \texttt{sim}\left( \mathcal{M}_{\text{emb}}(P_{\text{task}}), \mathcal{M}_{\text{emb}}(P^*_{\text{task}}) \right) < \delta \\
known & , \texttt{sim}\left( \mathcal{M}_{\text{emb}}(P_{\text{task}}), \mathcal{M}_{\text{emb}}(P^*_{\text{task}}) \right) \geq \delta
\end{cases}
\end{equation}
\end{small}

where $\texttt{sim}(*)$ is a cosine similarity function that computes the similarity between two embedding vectors.

By efficiently evaluating cognitive results, \textit{CPM} integrates alternative aggregation strategies for $\mathcal{M_{LLM}}$ based on $S_{cog}$ \footnote{we use $S_{cog}$ to conduct reverse evaluation of \texttt{GPT-4} in our experiments.}, as detailed in appendix~\ref{appcpm}: \textit{\textbf{(1) Solution logic aggregation for known tasks:}} $\mathcal{M_{LLM}}$ merges beneficial aspects from the original prompt $P$ with the LLM-taste prompt $P^*$ to create the final prompt $P_{\textit{final}}$. \textit{\textbf{(2) Stylistic template aggregation for unknown tasks}}:$\mathcal{M_{LLM}}$ extracts a cognitive preference template for thinking $T$ from the incorrect context in the LLM-taste prompt, and integrates meta-cognitive elements from the original prompt $P$ into $T$ to construct the final prompt $P_{final}$.

Consequently, we utilize the final LLM-preferred prompt to query the LLM $\mathcal{M_{LLM}}$ with a specific problem input, $problem$, to obtain the final logical answer, $answer$:
\begin{equation}
\label{query} answer = \mathcal{M_{LLM}}(P_{final}, problem)
\end{equation}

\section{Experiments}
\subsection{Datasets and Tasks}
To comprehensively validate our Reverse of Thought (RoT), following ToT~\cite{yao2024tree}, meta-prompting~\cite{suzgun2024meta} and BoT~\cite{yang2024buffer}, we assess it with baselines across a broad range of eight tasks across five logical benchmarks that encompass mathematical and algorithmic reasoning, domain-specific knowledge, and literary creativity: \textbf{(1) \texttt{Game of 24}} \cite{yao2024tree,xiang2025towards} challenges LLMs to create a mathematical expression utilizing each of four given numbers exactly once to achieve 24 \footnote{\url{https://www.4nums.com/game/}
}. We use 1000 samples drawn from a mix of both datasets for robust large-scale evaluation. 
\textbf{(2) \texttt{BIG-Bench} }\cite{suzgun2023challenging,srivastava2023beyond} involves \textbf{Geometric Shapes}, \textbf{Multi-Step Arithmetic Two} and \textbf{Word Sorting} and \textbf{Checkmate-in-One} from BIG-Bench suite \footnote{\url{https://huggingface.co/datasets/google/bigbench}} \cite{srivastava2023beyond}. \textbf{(3) \texttt{Python Puzzles}} \cite{schuster2021programming} comprises a collection of challenging programming puzzles crafted in Python, covering various difficulty levels \footnote{\url{https://github.com/microsoft/PythonProgrammingPuzzles}}. \textbf{(4) \texttt{Multilingual Grade School} Math (MGSM)} \cite{shi2022language} is a multilingual adaptation of the \texttt{GSM8K} dataset \cite{cobbe2021training}, featuring translations of a subset of examples into ten diverse languages\footnote{\url{https://github.com/google-research/url-nlp/tree/main/mgsm}}. \textbf{(5) \texttt{Shakespearean Sonnet Writing}} \cite{suzgun2024meta} require LLMs to compose with the rhyme scheme $\texttt{"ABAB CDCD EFEF GG"}$ while incorporating three \footnote{\url{https://huggingface.co/datasets/turingmachine/meta-prompting}} or five \footnote{\url{https://github.com/iljones00/Shakespearean-Sonnets-GPT}} specified words verbatim. 
\subsection{Baselines}
In our experiments, we compare RoT with five classic and latest state-of-the-art prompting baselines: 

\begin{itemize} 
    \setlength{\itemsep}{0pt}
    \item \textbf{CoT Prompting:} Following \citet{suzgun2024meta,yang2024buffer}, we employ \texttt{GPT-4} to decompose instruction into logic intermediate reasoning steps activated by \textit{"Let's think step by step"}.
    \item \textbf{Meta-Prompting:} \citet{suzgun2024meta} introduced general, task-agnostic prompts as a scaffold to guide LLMs effectively perform logic tasks.
    \item \textbf{Graph-of-Thought (GoT):} \citet{besta2024graph} modeled reasoning as a graph, where nodes are thoughts and edges define their relationships to solve complex problems.
    \item \textbf{Tree-of-Thought (ToT):} \citet{yao2024tree} organized reasoning in a tree structure, allowing LLMs to explore multiple thought paths and select the most promising ones for problem-solving, enhancing their complexity management in logical reasoning tasks.
    \item \textbf{Buffer of Thought (BoT):} \citet{yang2024buffer} introduced a meta-buffer that stores high-level thought templates, allowing for the adaptive retrieval and instantiation of relevant templates to enhance reasoning efficiency.
\end{itemize}
\subsection{Experiment Setup}
To ensure fair comparisons with previous methods, following \citet{yao2024tree,zhangcumulative,suzgun2024meta}, we utilize two LLMs: \texttt{GPT-3.5-turbo} and \texttt{GPT-4}, as the foundational model via OpenAI API for our RoT experiments including both main experiments and ablation study. For candidate generation, we use a temperature of 0.7, encouraging diversity in proposed solution. For instantiating reasoning steps, we adopt a lower temperature of 0.1, promoting more deterministic and stable logical progression. For the $warmup$ hyperparameter, we experimented with values of $\{1, 3, 5, 10\}$, accessed in batches for reverse reasoning warm-up through OpenAI API. Our findings suggest that a value of 5 optimally balances between logical accuracy and cost-efficiency. For embedding model $\mathcal{M}_{emb}$, we utilize a huggingface model \texttt{stella\_en\_1.5B\_v5} \footnote{\url{https://huggingface.co/dunzhang/stella_en_1.5B_v5}, \texttt{License:mit}}, a high-performance model with the smallest parameter countamong the top three on the \texttt{MTEB} leaderboard \cite{muennighoff2023mteb}, offline deployed on a single \texttt{NVIDIA GeForce RTX 4090 GPU}.

\paragraph{Knowledge Boundary and Cognitive Preference Consistency.}As described in Section~\ref{subsec:cognitive}, we trigger knowledge boundary using $S_{cog}$ by comparing $\texttt{sim}(P_{task},\mathcal{M_{LLM}}(D,P_r)_{task})$ with a threshold $\delta = 0.7$. We first evaluate \texttt{GPT-3.5-turbo} \& \texttt{GPT-4} on experimental tasks to distinguish between known and unknown domains in both 1-shot and 2-shot settings\footnote{To achieve better cost savings and fairness, we set the few-shot for all methods to 1-shot and 2-shot.} and subsequently conduct human evaluations of average cognitive preference consistency $\textit{Con}_{cog}$ based on stylistic and grammatical norms on $P^*$ across various tasks, with three professional annotators validating cognitive preferences on a 1–5 scale.

\subsection{Evaluation Metrics}
\begin{table*}[h]
\centering
\renewcommand{\arraystretch}{1.26}
\resizebox{0.98\textwidth}{!}{ 
\begin{tabular}{cc|ccccccccccc}
\toprule
 \multicolumn{2}{c}{Baselines} 
& \parbox{1.8cm}{\centering Game of 24} 
& \parbox{2cm}{\centering Geometric Shapes}
& \parbox{2cm}{\centering Multi-Step Arithmetic}
& \parbox{2cm}{\centering Word Sorting}
& \parbox{2cm}{\centering Checkmate-in-One}
& \parbox{2cm}{\centering Python Puzzles}
& \parbox{1.5cm}{\centering MGSM}
& \parbox{2cm}{\centering Sonnet Writing} \\
\midrule
\multicolumn{10}{c}{\texttt{\textbf{GPT-3.5-turbo}}} \\
\midrule
\multirow{2}{*}{\parbox{3.2cm}{\centering CoT (\textit{NeurIPS}) \\ \citet{wei2022chain} }} 
& 1-shot & 26.3 & 50.2 & 73.9 & 71.9 & 25.7 & 30.1 & 75.0 & 58.6 \\
\cline{2-2} 
& 2-shot & 29.6 & 53.2 & 76.2 & 74.6 & 25.4 & 30.6 & 77.4 & 61.3 \\

\midrule
\multirow{2}{*}{\parbox{3.2cm}{\centering Meta-prompting \citet{suzgun2024meta}}} 
& 1-shot & 59.6 & 69.5 & 83.1 & 90.0 & 47.4 & 40.2 & 80.3 & 69.5 \\
\cline{2-2} 
& 2-shot & 60.2 & 71.0 & 85.1 & 91.5 & 48.0 & 41.1 & 81.5 & 71.4 \\

\midrule
\multirow{2}{*}{\parbox{3.2cm}{\centering ToT (\textit{NeurIPS})  \citet{yao2024tree}}} 
& 1-shot & 66.7 & 56.2 & 81.9 & 89.2 & 44.2 & 38.2 & 78.3 & 63.5 \\
\cline{2-2} 
& 2-shot & 67.3 & 58.5 & 82.3 & 90.3 & 45.2 & 39.5 & 79.4 & 65.3 \\

\midrule
\multirow{2}{*}{\parbox{3.2cm}{\centering GoT (\textit{AAAI}) \\ \citet{besta2024graph}}} 
& 1-shot & 69.7 & 58.6 & 79.1 & 88.8 & 45.4 & 35.3 & 80.3 & 62.2 \\
\cline{2-2} 
& 2-shot & 70.4 & 58.9 & 79.0 & 89.5 & 46.0 & 36.7 & 81.0 & 63.7 \\

\midrule
\multirow{2}{*}{\parbox{3.2cm}{\centering BoT (\textit{NeurIPS}) \citet{yang2024buffer}}} 
& 1-shot & 74.7 & 83.1 & 87.6 & 93.2 & 67.9 & 48.2 & 81.1 & 71.1 \\
\cline{2-2} 
& 2-shot & 75.5 & 84.3 & 88.3 & 94.4 & 68.5 & 48.0 & 82.3 & 73.0 \\

\midrule
\multirow{2}{*}{\parbox{3.2cm}{\centering RoT (Ours)}} 
& 1-shot & \textbf{82.8} & \textbf{88.4} & \textbf{89.2} & \textbf{95.2} & \textbf{71.5} & \textbf{50.2} & \textbf{82.3} & \textbf{75.5} \\
\cline{2-2} 
& 2-shot & \textbf{87.8} & \textbf{88.7} & \textbf{89.5} & \textbf{95.6} & \textbf{72.6} & \textbf{50.4} & \textbf{84.7} & \textbf{75.8} \\

\midrule
\multicolumn{10}{c}{\texttt{\textbf{GPT-4}}} \\
\midrule
\multirow{2}{*}{\parbox{3.2cm}{\centering CoT (\textit{NeurIPS}) \\ \citet{wei2022chain} }} 
& 1-shot & 31.3 & 57.4 & 83.1 & 80.7 & 35.8 & 35.3 & 83.9 & 66.3 \\
\cline{2-2} 
& 2-shot & 32.7 & 60.9 & 85.5 & 82.7 & 36.7 & 35.9 & 84.3 & 70.2 \\

\midrule
\multirow{2}{*}{\parbox{3.2cm}{\centering Meta-prompting \citet{suzgun2024meta}}} 
& 1-shot & 64.6 & 76.7 & 89.6 & 97.6 & 58.2 & 45.0 & 85.1 & 78.7 \\
\cline{2-2} 
& 2-shot & 66.3 & 78.2 & 90.7 & 98.4 & 58.9 & 45.6 & 85.5 & 79.8 \\

\midrule
\multirow{2}{*}{\parbox{3.2cm}{\centering ToT (\textit{NeurIPS})  \citet{yao2024tree}}} 
& 1-shot & 73.7 & 58.6 & 88.8 & 95.2 & 49.8 & 43.8 & 84.7 & 68.7 \\
\cline{2-2} 
& 2-shot & 74.5 & 60.1 & 90.3 & 96.0 & 48.8 & 45.2 & 86.3 & 69.8 \\

\midrule
\multirow{2}{*}{\parbox{3.2cm}{\centering GoT (\textit{AAAI}) \\ \citet{besta2024graph}}} 
& 1-shot & 74.7 & 56.2 & 87.6 & 95.6 & 49.0 & 42.6 & 85.9 & 68.3 \\
\cline{2-2} 
& 2-shot & 73.5 & 57.7 & 88.7 & 96.8 & 51.2 & 43.5 & 86.7 & 62.5 \\

\midrule
\multirow{2}{*}{\parbox{3.2cm}{\centering BoT (\textit{NeurIPS}) \citet{yang2024buffer}}} 
& 1-shot & 82.8 & 90.4 & 94.8 & 99.2 & 87.1 & 51.8 & 87.9 & 79.1 \\
\cline{2-2} 
& 2-shot & 83.7 & 90.7 & 96.8 & 99.6 & 88.3 & 52.8 & 87.5 & 79.8 \\

\midrule
\multirow{2}{*}{\parbox{3.2cm}{\centering RoT (Ours)}} 
& 1-shot & \textbf{97.0} & \textbf{94.8} & \textbf{98.4} & \textbf{99.6} & \textbf{91.2} & \textbf{54.6} & \textbf{88.7} & \textbf{89.2} \\
\cline{2-2} 
& 2-shot & \textbf{98.0} & \textbf{95.2} & \textbf{99.2} & \textbf{100.0} & \textbf{92.0} & \textbf{57.3} & \textbf{90.0} & \textbf{92.0} \\
\bottomrule
\end{tabular}
}
\caption{Comparison of RoT with baselines across 1-shot and 2-shot settings for reasoning accuracy.}
\label{tab:accuracy}
\end{table*}
\paragraph{Reasoning Accuracy.}Following \citet{suzgun2024meta,yao2024tree}, we introduce a LLM\footnote{we select the latest openai-o1 as judger during the development of our work, \url{https://openai.com/index/introducing-openai-o1-preview/}} to validate the final logical reasoning against gold results (Correct/Wrong). We then compute logical accuracy $Acc_{logic}$ for each logical task by tallying the number of correct responses.

\paragraph{Reasoning Efficiency.}
Following \citet{yang2024buffer}, we evaluate reasoning efficiency in terms of complexity by calculating $T$, calculated as the average time spent per task across all samples in the test dataset:
\begin{equation}
\small
\label{time}
T = \frac{1}{N} \sum_{i=1}^{M} \sum_{j=1}^{N} T_{ij}
\end{equation}
where $N$ is the total number of tasks, $M$ is the total number of samples, and $T_{ij}$ represents the time taken for the $i$-th task on the $j$-th sample.
\section{Results and Discussion}
\subsection{Knowledge Boundary and Cognitive Preference Consistency}

\paragraph{Knowledge Boundary.}As shown in Table~\ref{knowledge}, our findings provide preliminary support for our hypothesis concerning cognitive knowledge boundaries of LLMs in reverse reasoning across current task benchmarks in one-shot and two-shot settings \cite{yao2024tree,suzgun2023challenging,srivastava2023beyond,schuster2021programming,shi2022language,suzgun2024meta}. Notably, \texttt{GPT-3.5-turbo \& GPT-4} excel in structured reasoning tasks such as \texttt{Game of 24}, \texttt{Geometric Shapes}, and \texttt{Checkmate-in-One}, consistently achieving strong results in both one-shot and two-shot settings. In contrast, tasks such as \texttt{MGSM (avg)} and \texttt{Python Puzzles} fall into an unknown category in the one-shot scenario, stemming from multi-source problems that are challenging to make reverse reasoning without sufficient context, indirectly supporting the rationale for our \textit{CPM} approach.

\paragraph{Cognitive Preference Consistency.}Our human evaluation of $\textit{Con}_{cog}$ yielded high scores of \textbf{4.32} in the 2-shot setting and \textbf{4.01} in the 1-shot setting, further validating that LLMs exhibit cognitive preferences shaped by RLHF \cite{ouyang2022training}. 

\begin{table}[ht]
\centering
\small
\begin{tabular}{lcc|lcc}
\toprule
\multicolumn{1}{c}{\multirow{2}{*}{Task}} & \multicolumn{2}{c|}{1-shot} & \multicolumn{2}{c}{2-shot} \\
\cmidrule(lr){2-3} \cmidrule(lr){4-5}
& Kno. & Unkno. & Kno. & Unkno. \\
\midrule
Game of 24            & \checkmark &   & \checkmark &   \\
MGSM (avg)            & &  \checkmark   & \checkmark &   \\
Multi-Step Arithmetic & \checkmark &   & \checkmark &   \\
WordSorting           & \checkmark &   & \checkmark &   \\
Python Puzzles        &            & \checkmark &  \checkmark          &  \\
Geometric Shapes      & \checkmark &   & \checkmark &   \\
Checkmate-in-One      & \checkmark &   & \checkmark &   \\
Sonnet Writing        & \checkmark &   & \checkmark &   \\
\bottomrule
\end{tabular}
\caption{Knowledge Boundary under Reversal Reasoning for \texttt{GPT-3.5-turbo} and \texttt{GPT-4} with identical results. \textit{Kno.} for Known, \textit{Unkno.} for Unknown.}
\setcounter{table}{1}
\label{knowledge}
\end{table}
\begin{table*}[h]
\large
\centering
\renewcommand{\arraystretch}{1}
\resizebox{0.98\textwidth}{!}{ 
\begin{tabular}{cc|ccccccccccc}
\toprule
 \multicolumn{2}{c}{Ablation study} 
& \parbox{1.8cm}{\centering Game of 24\textcolor{green}{\checkmark}} 
& \parbox{2.2cm}{\centering Geometric Shapes\textcolor{green}{\checkmark}}
& \parbox{2.4cm}{\centering Multi-Step Arithmetic\textcolor{green}{\checkmark}}
& \parbox{2cm}{\centering Word Sorting\textcolor{green}{\checkmark}}
& \parbox{2.3cm}{\centering Checkmate-in-One\textcolor{green}{\checkmark}}
& \parbox{2cm}{\centering Python Puzzles\textcolor{red}{$\times$}}
& \parbox{1.5cm}{\centering MGSM\textcolor{red}{$\times$}}
& \parbox{2cm}{\centering Sonnet Writing\textcolor{green}{\checkmark}} \\
\midrule
\multicolumn{10}{c}{\texttt{\textbf{GPT-3.5-turbo}}} \\

\midrule
\multirow{2}{*}{\parbox{3.2cm}{\centering \textbf{RoT}}} 
& 1-shot & \textbf{82.8} & \textbf{88.4} & \textbf{89.2} & \textbf{95.2} & \textbf{71.5} & \textbf{50.2} & \textbf{82.3} & \textbf{75.5} \\
\cline{2-2} 
& 2-shot & \textbf{87.8} & \textbf{88.7} & \textbf{89.5} & \textbf{95.6} & \textbf{72.6} & \textbf{50.4} & \textbf{84.7} & \textbf{75.8} \\

\midrule
\multirow{2}{*}{\parbox{3.2cm}{\centering \textbf{w/o \textit{PGRR}}}} 
& 1-shot & 74.7 & 72.3 & 74.3 & 74.7 & 65.5 & 44.2 & 78.7 & 69.6 \\
\cline{2-2} 
& 2-shot & 75.5 & 72.6 & 75.4 & 76.2 & 66.9 & 44.4 & 79.0 & 69.8 \\

\midrule
\multirow{2}{*}{\parbox{3.2cm}{\centering \textbf{w/o \textit{Logic}}}} 
& 1-shot & 76.7 & 80.7 & 81.5 & 84.3 & 66.3 & 45.8 & 77.5 & 72.9 \\
\cline{2-2} 
& 2-shot & 76.5 & 80.2 & 81.9 & 85.1 & 66.5 & 46.0 & 78.2 & 73.0 \\

\midrule
\multirow{2}{*}{\parbox{3.2cm}{\centering \textbf{w/o \textit{CPM}}}} 
& 1-shot & 80.8 & 86.0 & 88.4 & 93.2 & 70.3 & 39.8 & 72.7 & 74.2 \\
\cline{2-2} 
& 2-shot & 81.6 & 86.7 & 88.3 & 93.5 & 71.0 & 47.2 & 80.2 & 75.0 \\
\midrule
\multicolumn{10}{c}{\texttt{\textbf{GPT-4}}} \\
\midrule
\multirow{2}{*}{\parbox{3.2cm}{\centering \textbf{RoT}}} 
& 1-shot & \textbf{97.0} & \textbf{94.8} & \textbf{98.4} & \textbf{99.6} & \textbf{91.2} & \textbf{54.6} & \textbf{88.7} & \textbf{89.2} \\
\cline{2-2} 
& 2-shot & \textbf{98.0} & \textbf{95.2} & \textbf{99.2} & \textbf{100.0} & \textbf{92.0} & \textbf{57.3} & \textbf{90.0} & \textbf{92.0} \\

\midrule
\multirow{2}{*}{\parbox{3.2cm}{\centering \textbf{w/o \textit{PGRR}}}} 
& 1-shot & 77.8 & 74.7 & 78.7 & 79.5 & 80.7 & 45.0 & 85.9 & 72.7 \\
\cline{2-2} 
& 2-shot & 78.6 & 75.0 & 79.0 & 81.0 & 82.3 & 46.8 & 86.0 & 73.0 \\

\midrule
\multirow{2}{*}{\parbox{3.2cm}{\centering \textbf{w/o \textit{Logic}}}} 
& 1-shot & 85.9 & 83.1 & 87.1 & 89.2 & 81.5 & 49.8 & 85.5 & 86.1 \\
\cline{2-2} 
& 2-shot & 86.7 & 83.5 & 87.9 & 89.5 & 83.5 & 50.0 & 86.7 & 77.0 \\

\midrule
\multirow{2}{*}{\parbox{3.2cm}{\centering\textbf{ w/o \textit{CPM}}}} 
& 1-shot & 92.0 & 90.0 & 94.8 & 97.6 & 89.2 & 40.6 & 84.7 & 88.8 \\
\cline{2-2} 
& 2-shot & 93.9 & 90.3 & 95.2 & 98.4 & 89.1 & 51.6 & 86.3 & 91.5 \\
\bottomrule
\end{tabular}
}
\caption{Ablation study of RoT across various tasks in 1-shot and 2-shot settings. \textcolor{green}{\checkmark} indicates known tasks in both 1-shot and 2-shot; \textcolor{red}{$\times$} indicates unknown tasks in 1-shot.}
\label{tab:ablation}
\end{table*}
\vspace{-0.5cm}
\subsection{Reasoning Accuracy and Efficiency}
\paragraph{(1) RoT can activate LLMs' reasoning accuracy.}
As shown in Table~\ref{tab:accuracy}, RoT consistently outperforms all baselines across various reasoning tasks, with particularly notable improvements observed in \texttt{GPT-4}. Specifically, compared with the best BoT, RoT achieves significant gains in tasks such as \texttt{Game of 24} \textbf{(+17.15\% in 1-shot and +17.08\% in 2-shot)}, \texttt{Geometric Shapes} \textbf{(+4.87\% in 1-shot and +4.96\% in 2-shot)}, and \texttt{Checkmate-in-One} \textbf{(+4.71\% in 1-shot and +4.19\% in 2-shot)}, demonstrating our flexibility  of leveraging cognitive preference in 
LLM to activate logic capabilities through reverse reasoning (case study detailed in appendix~\ref{sec:case}). For \texttt{GPT-3.5-turbo}, RoT also demonstrates substantial improvements, such as \texttt{Game of 24} \textbf{(+10.84\% in 1-shot and +16.29\% in 2-shot)} and \texttt{Geometric Shapes} \textbf{(+6.38\% in 1-shot and +5.22\% in 2-shot)}, further emphasizing its versatility in activating logic through cognitive preference management.

\begin{figure}[ht]
  \includegraphics[width=\linewidth]{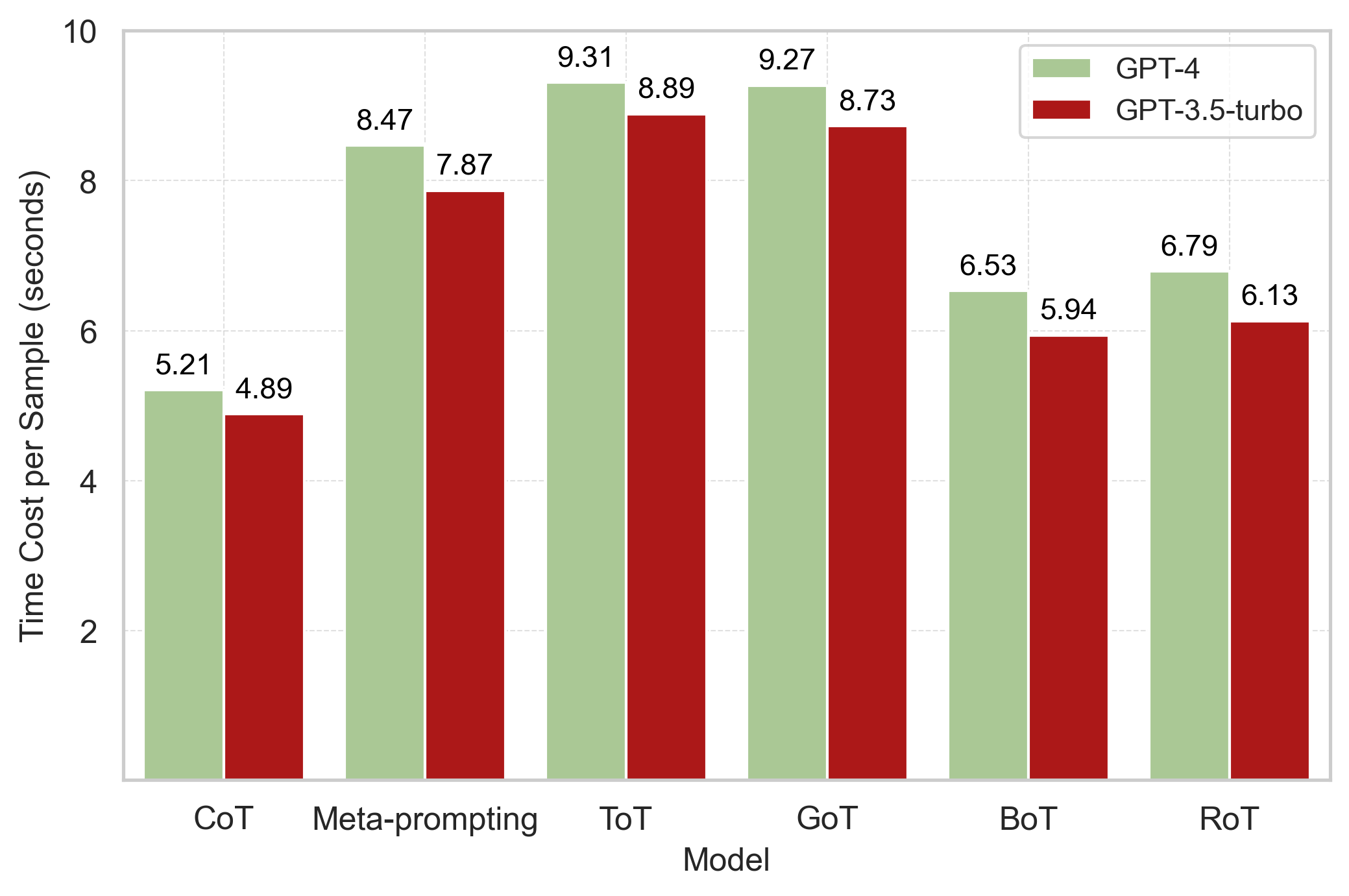}
\caption{Inference time comparison, measured as the average duration from inference start to evaluation end, including all steps.}
  \label{fig:cost}
\end{figure}

\paragraph{(2) RoT demonstrates better tradeoff between reasoning accuracy and efficiency across baselines.}
As shown in Figure~\ref{fig:cost}, RoT achieves competitive performance in reasoning efficiency, outperforming baselines and being second only to BoT, while BoT’s dependence on numerous pre-defined golden thought templates limits its flexibility. In contrast, our RoT, as a plugin strategy,  emphasizes exploring optimal prompt for solution and task data instantiation after reverse reasoning warm-up.

\subsection{Ablation Study}
\label{ablation}

As shown in Table \ref{tab:ablation}, we conducted three ablation studies to evaluate key components: (1) \textbf{w/o \textit{PGRR}}: Removing Preference-Guided Reverse Reasoning(\textit{PGRR}) for exploring LLMs-taste prompts; (2) \textbf{w/o \textit{Logic}}: Excluding mathematical logic for pseudo-code plan-and-solve; and (3) \textbf{w/o \textit{CPM}}: Eliminating Cognitive Preference Manage(\textit{CPM}) for both known and unknown tasks.
\vspace{-0.3em}
\paragraph{Impact of \textit{PGRR}.}
Excluding Preference-Guided Reverse Reasoning (\textit{PGRR}) results in a significant reduction in overall task performance, as seen in tasks like the Game of 24 (98.0\% to 78.6\% for \texttt{GPT-4}, 87.8\% to 75.5\% for \texttt{GPT-3.5-turbo}) and WordSorting (100\% to 81\% for \texttt{GPT-4}, 95.6\% to 76.2\% for \texttt{GPT-3.5-turbo}), indicating \textbf{w/o \textit{PGRR}} weaken model's cognition to task-specific requirements.
\vspace{-0.3em}
\paragraph{Impact of \textit{Logic}.}
Removing mathematical logic from RoT leads to notable declines in tasks requiring structured problem-solving, such as Multi-Step Arithmetic (99.2\% to 87.9\% for \texttt{GPT-4}, 89.5\% to 81.9\% for \texttt{GPT-3.5-turbo}) and Checkmate-in-One (92.0\% to 83.5\% for \texttt{GPT-4}, 72.6\% to 66.5\% for \texttt{GPT-3.5-turbo}), underscoring \textbf{w/o \textit{Logic}} negatively affects structured problem-solving for complex reasoning.
\paragraph{Impact of \textit{CPM}.}
Lacking Cognitive Preference Manager (\textit{CPM}) has a particularly pronounced effect on unknown tasks, with Python Puzzles dropping from 54.6\% to 45.0\% for \texttt{GPT-4} and 50.2\% to 39.8\% for \texttt{GPT-3.5-turbo}, while MGSM decreased from 84.7\% to 80.2\% for \texttt{GPT-3.5-turbo} and from 90.0\% to 86.3\% for \texttt{GPT-4}, indicating \textbf{w/o \textit{CPM}} weaken RoT's flexibility for tackling known tasks and unknown tasks \footnote{Please refer to Appendix~\ref{sec:knowncase} for case study on known task and Appendix~\ref{sec:unknowncase} for case study on unknwon task.}. 

\section{Conclusion}
In this paper, we propose Reversal of Thought (RoT), a novel and plug-and-play framework to enhance the logical reasoning capabilities of LLMs. By integrating reverse reasoning with meta-cognitive mechanisms and cognitive preference management, RoT improves reasoning accuracy and efficiency while minimizing computational costs, which leverages \textit{Preference-Guided Reverse Reasoning} and \textit{Cognitive Preference Manager}, which optimally aligns LLM reasoning processes with their cognitive preferences shaped by their pretraining and RLHF. Comprehensive experiments across diverse reasoning tasks demonstrate that RoT consistently outperforms state-of-the-art baselines in both known and unknown task scenarios, demonstrating the potential to expand knowledge boundaries through cognitive preference template. Our research provides valuable insights into future studies focused on further enhancing LLMs’ reasoning capacities by dynamically exploring cognitive preferences for complex reasoning tasks.

\section*{Limitations}
Reversal of Thought (RoT) introduce a reverse reasoning warm-up to activate cognitive preference for LLMs to enhance logic capabilities and introduce a cognitive preference manager to determine knowledge boundary and utilize cognitive preference for known and unknown tasks. 

While RoT has performed exceptionally in logic accuracy and efficiency, We discuss major challenge in its reliance on two-shot demonstration inputs involving two distinct problem cases. We observed that RoT may struggles with one-shot learning in multi-source tasks. we partially and effectively mitigates this issue through the integration of \textit{Cognitive Preference Manager (CPM)} and two-shot learning. 

In future work, we aim to extend RoT’s capabilities by incorporating In-Context Learning (ICL) or, alternatively, adopting Auto-prompt based demonstration selection strategies \cite{zhangautomatic,jin2024self}, which will allow for greater flexibility in adapting to varied contexts and improve its performance on more complex reasoning tasks. Furthermore, we believe that utilizing Reversal of Thought in teacher-student model distillation, which could further amplify the practical value of our approach. 

Finally, we acknowledge the challenges posed by the inherent instability of large language models. Future research may benefit from investigating robust candidate solutions—such as hybrid scoring mechanisms \cite{wang2024integrate,minh2024turning,wang2024conu}—that can be integrated with RoT to further enhance its reliability and overall performance through joint optimization.

\section*{Ethical Considerations}
Since the datasets used in our experiments focus on pure mathematical and algorithmic reasoning, domain-specific knowledge, and literary creativity, which are all sourced from publicly available datasets \cite{yao2024tree,suzgun2023challenging,srivastava2023beyond,schuster2021programming,shi2022language,suzgun2024meta,xiang2025towards} and devoid of any personal privacy or sensitive ethical information. Therefore, we do not identify any immediate ethical concerns regarding our current work. Additionally, we conduct human evaluations of cognitive preference consistency with three anonymous professional annotators following our instructions (detailed in appendix~\ref{sec:instruction}).

\bibliography{custom}

\begin{thebibliography}{47}
\providecommand{\natexlab}[1]{#1}

\bibitem[{Achiam et~al.(2023)Achiam, Adler, Agarwal, Ahmad, Akkaya, Aleman, Almeida, Altenschmidt, Altman, Anadkat et~al.}]{achiam2023gpt}
Josh Achiam, Steven Adler, Sandhini Agarwal, Lama Ahmad, Ilge Akkaya, Florencia~Leoni Aleman, Diogo Almeida, Janko Altenschmidt, Sam Altman, Shyamal Anadkat, et~al. 2023.
\newblock Gpt-4 technical report.
\newblock \emph{arXiv preprint arXiv:2303.08774}.

\bibitem[{Arkoudas(2023)}]{arkoudas2023gpt}
Konstantine Arkoudas. 2023.
\newblock Gpt-4 can't reason.
\newblock \emph{arXiv preprint arXiv:2308.03762}.

\bibitem[{Bai et~al.(2023)Bai, Bai, Chu, Cui, Dang, Deng, Fan, Ge, Han, Huang et~al.}]{bai2023qwen}
Jinze Bai, Shuai Bai, Yunfei Chu, Zeyu Cui, Kai Dang, Xiaodong Deng, Yang Fan, Wenbin Ge, Yu~Han, Fei Huang, et~al. 2023.
\newblock Qwen technical report.
\newblock \emph{arXiv preprint arXiv:2309.16609}.

\bibitem[{Bao et~al.(2024)Bao, Zhang, Yang, Wang, and Zhang}]{bao2024llms}
Guangsheng Bao, Hongbo Zhang, Linyi Yang, Cunxiang Wang, and Yue Zhang. 2024.
\newblock Llms with chain-of-thought are non-causal reasoners.
\newblock \emph{arXiv preprint arXiv:2402.16048}.

\bibitem[{Besta et~al.(2024)Besta, Blach, Kubicek, Gerstenberger, Podstawski, Gianinazzi, Gajda, Lehmann, Niewiadomski, Nyczyk et~al.}]{besta2024graph}
Maciej Besta, Nils Blach, Ales Kubicek, Robert Gerstenberger, Michal Podstawski, Lukas Gianinazzi, Joanna Gajda, Tomasz Lehmann, Hubert Niewiadomski, Piotr Nyczyk, et~al. 2024.
\newblock Graph of thoughts: Solving elaborate problems with large language models.
\newblock In \emph{Proceedings of the AAAI Conference on Artificial Intelligence}, volume~38, pages 17682--17690.

\bibitem[{Bi et~al.(2025)Bi, Yan, Wang, Huang, Chen, Wan, Ye, Xiao, Schuetze, Tresp, and Ma}]{bi2025cotkinetics}
Jinhe Bi, Danqi Yan, Yifan Wang, Wenke Huang, Haokun Chen, Guancheng Wan, Mang Ye, Xun Xiao, Hinrich Schuetze, Volker Tresp, and Yunpu Ma. 2025.
\newblock \href {https://arxiv.org/abs/2505.13408} {Cot-kinetics: A theoretical modeling assessing lrm reasoning process}.
\newblock \emph{Preprint}, arXiv:2505.13408.

\bibitem[{Cai et~al.(2023)Cai, Wang, Ma, Chen, and Zhou}]{cai2023large}
Tianle Cai, Xuezhi Wang, Tengyu Ma, Xinyun Chen, and Denny Zhou. 2023.
\newblock Large language models as tool makers.
\newblock \emph{arXiv preprint arXiv:2305.17126}.

\bibitem[{Chen et~al.()Chen, Qin, Jiaqi, Zhou, and Che}]{chenunlocking}
Qiguang Chen, Libo Qin, WANG Jiaqi, Jingxuan Zhou, and Wanxiang Che.
\newblock Unlocking the capabilities of thought: A reasoning boundary framework to quantify and optimize chain-of-thought.
\newblock In \emph{The Thirty-eighth Annual Conference on Neural Information Processing Systems}.

\bibitem[{Cobbe et~al.(2021)Cobbe, Kosaraju, Bavarian, Chen, Jun, Kaiser, Plappert, Tworek, Hilton, Nakano et~al.}]{cobbe2021training}
Karl Cobbe, Vineet Kosaraju, Mohammad Bavarian, Mark Chen, Heewoo Jun, Lukasz Kaiser, Matthias Plappert, Jerry Tworek, Jacob Hilton, Reiichiro Nakano, et~al. 2021.
\newblock Training verifiers to solve math word problems.
\newblock \emph{arXiv preprint arXiv:2110.14168}.

\bibitem[{Dubey et~al.(2024)Dubey, Jauhri, Pandey, Kadian, Al-Dahle, Letman, Mathur, Schelten, Yang, Fan et~al.}]{dubey2024llama}
Abhimanyu Dubey, Abhinav Jauhri, Abhinav Pandey, Abhishek Kadian, Ahmad Al-Dahle, Aiesha Letman, Akhil Mathur, Alan Schelten, Amy Yang, Angela Fan, et~al. 2024.
\newblock The llama 3 herd of models.
\newblock \emph{arXiv preprint arXiv:2407.21783}.

\bibitem[{Fleur et~al.(2021)Fleur, Bredeweg, and van~den Bos}]{fleur2021metacognition}
Damien~S Fleur, Bert Bredeweg, and Wouter van~den Bos. 2021.
\newblock Metacognition: ideas and insights from neuro-and educational sciences.
\newblock \emph{npj Science of Learning}, 6(1):13.

\bibitem[{Franklin(2005)}]{franklin2005tell}
Benjamin Franklin. 2005.
\newblock Tell me and i forget. teach me and i remember. involve me and i learn.

\bibitem[{Jin and Lu(2024)}]{jin2024self}
Ziqi Jin and Wei Lu. 2024.
\newblock Self-harmonized chain of thought.
\newblock \emph{arXiv preprint arXiv:2409.04057}.

\bibitem[{Lee and Kim(2023)}]{lee2023recursion}
Soochan Lee and Gunhee Kim. 2023.
\newblock Recursion of thought: A divide-and-conquer approach to multi-context reasoning with language models.
\newblock In \emph{Findings of the Association for Computational Linguistics: ACL 2023}, pages 623--658.

\bibitem[{Liu et~al.(2024)Liu, Zhou, Guo, Shareghi, Vulic, Korhonen, and Collier}]{liu2024aligning}
Yinhong Liu, Han Zhou, Zhijiang Guo, Ehsan Shareghi, Ivan Vulic, Anna Korhonen, and Nigel Collier. 2024.
\newblock Aligning with human judgement: The role of pairwise preference in large language model evaluators.
\newblock \emph{arXiv preprint arXiv:2403.16950}.

\bibitem[{Madaan et~al.(2024)Madaan, Tandon, Gupta, Hallinan, Gao, Wiegreffe, Alon, Dziri, Prabhumoye, Yang et~al.}]{madaan2024self}
Aman Madaan, Niket Tandon, Prakhar Gupta, Skyler Hallinan, Luyu Gao, Sarah Wiegreffe, Uri Alon, Nouha Dziri, Shrimai Prabhumoye, Yiming Yang, et~al. 2024.
\newblock Self-refine: Iterative refinement with self-feedback.
\newblock \emph{Advances in Neural Information Processing Systems}, 36.

\bibitem[{Margatina et~al.(2023)Margatina, Schick, Aletras, and Dwivedi-Yu}]{margatina2023active}
Katerina Margatina, Timo Schick, Nikolaos Aletras, and Jane Dwivedi-Yu. 2023.
\newblock Active learning principles for in-context learning with large language models.
\newblock In \emph{Findings of the Association for Computational Linguistics: EMNLP 2023}, pages 5011--5034.

\bibitem[{Minh et~al.(2024)Minh, Baker, Neo, Roush, Kirsch, Independent, Shwartz-Ziv, and AI}]{minh2024turning}
Nguyen~Nhat Minh, Andrew Baker, Clement Neo, Allen Roush, Andreas Kirsch, Wand~AI Independent, Ravid Shwartz-Ziv, and Wand AI. 2024.
\newblock Turning up the heat: Min-p sampling for creative and coherent llm outputs.
\newblock In \emph{The Thirteenth International Conference on Learning Representations}.

\bibitem[{Muennighoff et~al.(2023)Muennighoff, Tazi, Magne, and Reimers}]{muennighoff2023mteb}
Niklas Muennighoff, Nouamane Tazi, Loic Magne, and Nils Reimers. 2023.
\newblock Mteb: Massive text embedding benchmark.
\newblock In \emph{Proceedings of the 17th Conference of the European Chapter of the Association for Computational Linguistics}, pages 2014--2037.

\bibitem[{Narang et~al.()Narang, Chowdhery, and Zhou}]{narangself}
Sharan Narang, Aakanksha Chowdhery, and Denny Zhou.
\newblock Self-consistency improves chain of thought reasoning in language models.

\bibitem[{Ouyang et~al.(2022)Ouyang, Wu, Jiang, Almeida, Wainwright, Mishkin, Zhang, Agarwal, Slama, Ray et~al.}]{ouyang2022training}
Long Ouyang, Jeffrey Wu, Xu~Jiang, Diogo Almeida, Carroll Wainwright, Pamela Mishkin, Chong Zhang, Sandhini Agarwal, Katarina Slama, Alex Ray, et~al. 2022.
\newblock Training language models to follow instructions with human feedback.
\newblock \emph{Advances in neural information processing systems}, 35:27730--27744.

\bibitem[{Pan et~al.()Pan, Albalak, Wang, and Wang}]{pan2023logic}
Liangming Pan, Alon Albalak, Xinyi Wang, and William~Yang Wang.
\newblock Logic-lm: Empowering large language models with symbolic solvers for faithful logical reasoning.
\newblock In \emph{The 2023 Conference on Empirical Methods in Natural Language Processing}.

\bibitem[{Plaat et~al.(2024)Plaat, Wong, Verberne, Broekens, van Stein, and Back}]{plaat2024reasoning}
Aske Plaat, Annie Wong, Suzan Verberne, Joost Broekens, Niki van Stein, and Thomas Back. 2024.
\newblock Reasoning with large language models, a survey.
\newblock \emph{arXiv preprint arXiv:2407.11511}.

\bibitem[{Renze and Guven(2024)}]{renze2024self}
Matthew Renze and Erhan Guven. 2024.
\newblock Self-reflection in llm agents: Effects on problem-solving performance.
\newblock \emph{arXiv preprint arXiv:2405.06682}.

\bibitem[{Schuster et~al.(2021)Schuster, Kalyan, Polozov, and Kalai}]{schuster2021programming}
Tal Schuster, Ashwin Kalyan, Alex Polozov, and Adam~Tauman Kalai. 2021.
\newblock \href {https://arxiv.org/abs/2106.05784} {Programming puzzles}.
\newblock In \emph{Thirty-fifth Conference on Neural Information Processing Systems Datasets and Benchmarks Track}.

\bibitem[{Shi et~al.(2022)Shi, Suzgun, Freitag, Wang, Srivats, Vosoughi, Chung, Tay, Ruder, Zhou et~al.}]{shi2022language}
Freda Shi, Mirac Suzgun, Markus Freitag, Xuezhi Wang, Suraj Srivats, Soroush Vosoughi, Hyung~Won Chung, Yi~Tay, Sebastian Ruder, Denny Zhou, et~al. 2022.
\newblock Language models are multilingual chain-of-thought reasoners.
\newblock In \emph{The Eleventh International Conference on Learning Representations}.

\bibitem[{Srivastava et~al.(2023)Srivastava, Rastogi, Rao, Shoeb, Abid, Fisch, Brown, Santoro, Gupta, Garriga-Alonso et~al.}]{srivastava2023beyond}
Aarohi Srivastava, Abhinav Rastogi, Abhishek Rao, Abu Awal~Md Shoeb, Abubakar Abid, Adam Fisch, Adam~R Brown, Adam Santoro, Aditya Gupta, Adri{\`a} Garriga-Alonso, et~al. 2023.
\newblock Beyond the imitation game: Quantifying and extrapolating the capabilities of language models.
\newblock \emph{Transactions on Machine Learning Research}.

\bibitem[{Stechly et~al.()Stechly, Marquez, and Kambhampati}]{stechly2023gpt}
Kaya Stechly, Matthew Marquez, and Subbarao Kambhampati.
\newblock Gpt-4 doesn’t know it’s wrong: An analysis of iterative prompting for reasoning problems.
\newblock In \emph{NeurIPS 2023 Foundation Models for Decision Making Workshop}.

\bibitem[{Sun et~al.(2024)Sun, Xu, Liu, Luan, Wang, Shang, Wen, and Yan}]{sun2024determlr}
Hongda Sun, Weikai Xu, Wei Liu, Jian Luan, Bin Wang, Shuo Shang, Ji-Rong Wen, and Rui Yan. 2024.
\newblock Determlr: Augmenting llm-based logical reasoning from indeterminacy to determinacy.
\newblock In \emph{Proceedings of the 62nd Annual Meeting of the Association for Computational Linguistics (Volume 1: Long Papers)}, pages 9828--9862.

\bibitem[{Suzgun and Kalai(2024)}]{suzgun2024meta}
Mirac Suzgun and Adam~Tauman Kalai. 2024.
\newblock Meta-prompting: Enhancing language models with task-agnostic scaffolding.
\newblock \emph{arXiv preprint arXiv:2401.12954}.

\bibitem[{Suzgun et~al.(2023)Suzgun, Scales, Sch{\"a}rli, Gehrmann, Tay, Chung, Chowdhery, Le, Chi, Zhou et~al.}]{suzgun2023challenging}
Mirac Suzgun, Nathan Scales, Nathanael Sch{\"a}rli, Sebastian Gehrmann, Yi~Tay, Hyung~Won Chung, Aakanksha Chowdhery, Quoc Le, Ed~Chi, Denny Zhou, et~al. 2023.
\newblock Challenging big-bench tasks and whether chain-of-thought can solve them.
\newblock In \emph{Findings of the Association for Computational Linguistics: ACL 2023}, pages 13003--13051.

\bibitem[{Tian et~al.(2024)Tian, Xu, and Mao}]{tian2024theory}
Yuan Tian, Nan Xu, and Wenji Mao. 2024.
\newblock A theory guided scaffolding instruction framework for llm-enabled metaphor reasoning.
\newblock In \emph{Proceedings of the 2024 Conference of the North American Chapter of the Association for Computational Linguistics: Human Language Technologies (Volume 1: Long Papers)}, pages 7731--7748.

\bibitem[{Uddin(2021)}]{uddin2021cognitive}
Lucina~Q Uddin. 2021.
\newblock Cognitive and behavioural flexibility: neural mechanisms and clinical considerations.
\newblock \emph{Nature Reviews Neuroscience}, 22(3):167--179.

\bibitem[{Wang et~al.(2024{\natexlab{a}})Wang, Feng, Li, Yuan, Zhang, Pan, Wang, Hu, and Li}]{wang2024make}
Xinglin Wang, Shaoxiong Feng, Yiwei Li, Peiwen Yuan, Yueqi Zhang, Boyuan Pan, Heda Wang, Yao Hu, and Kan Li. 2024{\natexlab{a}}.
\newblock Make every penny count: Difficulty-adaptive self-consistency for cost-efficient reasoning.
\newblock \emph{arXiv preprint arXiv:2408.13457}.

\bibitem[{Wang et~al.(2024{\natexlab{b}})Wang, Li, Feng, Yuan, Pan, Wang, Hu, and Li}]{wang2024integrate}
Xinglin Wang, Yiwei Li, Shaoxiong Feng, Peiwen Yuan, Boyuan Pan, Heda Wang, Yao Hu, and Kan Li. 2024{\natexlab{b}}.
\newblock Integrate the essence and eliminate the dross: Fine-grained self-consistency for free-form language generation.
\newblock In \emph{Proceedings of the 62nd Annual Meeting of the Association for Computational Linguistics (Volume 1: Long Papers)}, pages 11782--11794.

\bibitem[{Wang et~al.(2024{\natexlab{c}})Wang, Duan, Cheng, Zhang, Wang, Shi, Xu, Shen, and Zhu}]{wang2024conu}
Zhiyuan Wang, Jinhao Duan, Lu~Cheng, Yue Zhang, Qingni Wang, Xiaoshuang Shi, Kaidi Xu, Heng~Tao Shen, and Xiaofeng Zhu. 2024{\natexlab{c}}.
\newblock Conu: Conformal uncertainty in large language models with correctness coverage guarantees.
\newblock In \emph{Findings of the Association for Computational Linguistics: EMNLP 2024}, pages 6886--6898.

\bibitem[{Wei et~al.(2022)Wei, Wang, Schuurmans, Bosma, Xia, Chi, Le, Zhou et~al.}]{wei2022chain}
Jason Wei, Xuezhi Wang, Dale Schuurmans, Maarten Bosma, Fei Xia, Ed~Chi, Quoc~V Le, Denny Zhou, et~al. 2022.
\newblock Chain-of-thought prompting elicits reasoning in large language models.
\newblock \emph{Advances in neural information processing systems}, 35:24824--24837.

\bibitem[{Xiang et~al.(2025)Xiang, Snell, Gandhi, Albalak, Singh, Blagden, Phung, Rafailov, Lile, Mahan et~al.}]{xiang2025towards}
Violet Xiang, Charlie Snell, Kanishk Gandhi, Alon Albalak, Anikait Singh, Chase Blagden, Duy Phung, Rafael Rafailov, Nathan Lile, Dakota Mahan, et~al. 2025.
\newblock Towards system 2 reasoning in llms: Learning how to think with meta chain-of-though.
\newblock \emph{arXiv preprint arXiv:2501.04682}.

\bibitem[{Yang et~al.(2024{\natexlab{a}})Yang, Yu, Zhang, Cao, Xu, Zhang, Gonzalez, and Cui}]{yang2024buffer}
Ling Yang, Zhaochen Yu, Tianjun Zhang, Shiyi Cao, Minkai Xu, Wentao Zhang, Joseph~E Gonzalez, and Bin Cui. 2024{\natexlab{a}}.
\newblock Buffer of thoughts: Thought-augmented reasoning with large language models.
\newblock \emph{Advances in Neural Information Processing Systems}, 37:113519--113544.

\bibitem[{Yang et~al.(2024{\natexlab{b}})Yang, Chen, and Tam}]{yang2024arithmetic}
Xiaocheng Yang, Bingsen Chen, and Yik-Cheung Tam. 2024{\natexlab{b}}.
\newblock Arithmetic reasoning with llm: Prolog generation \& permutation.
\newblock In \emph{Proceedings of the 2024 Conference of the North American Chapter of the Association for Computational Linguistics: Human Language Technologies (Volume 2: Short Papers)}, pages 699--710.

\bibitem[{Yao et~al.(2024)Yao, Yu, Zhao, Shafran, Griffiths, Cao, and Narasimhan}]{yao2024tree}
Shunyu Yao, Dian Yu, Jeffrey Zhao, Izhak Shafran, Tom Griffiths, Yuan Cao, and Karthik Narasimhan. 2024.
\newblock Tree of thoughts: Deliberate problem solving with large language models.
\newblock \emph{Advances in Neural Information Processing Systems}, 36.

\bibitem[{Yin et~al.(2024)Yin, Zhang, Ruan, and Wan}]{yin2024benchmarking}
Xunjian Yin, Xu~Zhang, Jie Ruan, and Xiaojun Wan. 2024.
\newblock Benchmarking knowledge boundary for large language model: A different perspective on model evaluation.
\newblock \emph{arXiv preprint arXiv:2402.11493}.

\bibitem[{Yu et~al.(2024)Yu, Peng, Galley, Gao, and Yu}]{yu2024teaching}
Xiao Yu, Baolin Peng, Michel Galley, Jianfeng Gao, and Zhou Yu. 2024.
\newblock Teaching language models to self-improve through interactive demonstrations.
\newblock In \emph{Proceedings of the 2024 Conference of the North American Chapter of the Association for Computational Linguistics: Human Language Technologies (Volume 1: Long Papers)}, pages 5127--5149.

\bibitem[{Zhang et~al.()Zhang, Yang, Yuan, and Yao}]{zhangcumulative}
Yifan Zhang, Jingqin Yang, Yang Yuan, and Andrew~C Yao.
\newblock Cumulative reasoning with large language models.

\bibitem[{Zhang et~al.(2023)Zhang, Zhang, Li, and Smola}]{zhangautomatic}
Zhuosheng Zhang, Aston Zhang, Mu~Li, and Alex Smola. 2023.
\newblock Automatic chain of thought prompting in large language models.
\newblock In \emph{The Eleventh International Conference on Learning Representations}.

\bibitem[{Zhou et~al.()Zhou, Sch{\"a}rli, Hou, Wei, Scales, Wang, Schuurmans, Cui, Bousquet, Le et~al.}]{zhouleast}
Denny Zhou, Nathanael Sch{\"a}rli, Le~Hou, Jason Wei, Nathan Scales, Xuezhi Wang, Dale Schuurmans, Claire Cui, Olivier Bousquet, Quoc~V Le, et~al.
\newblock Least-to-most prompting enables complex reasoning in large language models.
\newblock In \emph{The Eleventh International Conference on Learning Representations}.

\bibitem[{Zhou et~al.(2023)Zhou, Xu, He, Zhou, Zhai, Chen, Long, Zheng, and Lu}]{zhou2023biasing}
Siyuan Zhou, Xinran Xu, Xiangyu He, Faxin Zhou, Yu~Zhai, Jinglu Chen, Yuhang Long, Lifen Zheng, and Chunming Lu. 2023.
\newblock Biasing the neurocognitive processing of videos with the presence of a real cultural other.
\newblock \emph{Cerebral Cortex}, 33(4):1090--1103.

\end{thebibliography}
\appendix
\section{Experimental Details} 

\subsection{Prompt for Reverse Reasoning}
\label{subsec:prompt}
\begin{center}
\begin{tcolorbox}[width=\textwidth,colback=white, colframe=NavyBlue, title=Prompt for Reverse Reasoning]
\label{prompt}
{\slshape 
\textbf{Instruction}\\
You are a highly distinguished expert in mathematics and information reasoning. 
Based on the given example, define the specific task, including the task definition, pseudocode, logical pseudocode, case examples, and input-output format. \\
\textcolor{BlueGreen}{\textbf{1. Understand Task Description:}}\\ 
Meticulously study demonstrations to deeply understand generic task description.\\
\textcolor{Plum}{\textbf{2. Plan Generic 
 Pseudocode:}}\\ 
Provide pseudocode in text form and plan an efficient algorithm to complete the task with your experiences.\\
\textcolor{ForestGreen}{\textbf{3. Formulate Logical Pseudocode:}}\\ 
Convert the pseudocode into generic logical algorithm pseudocode using \textcolor{Red}{ONLY logical symbols}: \\
\textbf{Logical Operators}: \\ Conjunction: $A \land B$ ; Disjunction: $A \lor B$ \\
equivalence: $A \equiv B$ , Negation: $\neg A$ \\ 
\textbf{Quantifiers:}\\
Universal quantifier: $\forall x$; Existential quantifier: $\exists x$ \\
\textbf{Inequalities:} \\
Less than: $x < y$; Greater than: $x > y$ \\
Less than or equal to: $x \leq y$ \\
Greater than or equal to: $x \geq y$ \\
Equals: $x = y$; Not equals: $x \neq y$ \\
\textbf{Conditional Statements:} \\
If $A$ then $B$: $A \supset B$ \\
If $A \land B$ then $C$: $(A \land B) \supset C$ \\
If $A \lor B$ then $C$: $(A \lor B) \supset C$ \\
If $\forall x (P(x))$ then $Q$: $\forall x (P(x)) \supset Q$ \\
If $\exists x (P(x))$ then $Q$: $\exists x (P(x)) \supset Q$  etc. \\
\textbf{Input:} [Demonstration] \textbf{Output:} [Output]
}
\end{tcolorbox}
\noindent\begin{minipage}{\textwidth}
\captionof{figure}{Prompt for Reverse Reasoning}\label{prompt}
\end{minipage}
\end{center}

\clearpage
\subsection{Prompt for CPM (Known/Unknown)}
\label{appcpm}
\begin{center}
\begin{tcolorbox}[width=\textwidth,colback=white, colframe=NavyBlue, title=Prompt for CPM (Known/Unknown)]
\label{promptagg}
{\slshape 
\textbf{Instruction}\\
You are an expert in information synthesis, proficient in combining complementary insights and extracting essential details from the viewpoints of the distilled task definition, detailed generic logical pseudocode, case example, and input-output format.  \\
\textcolor{BlueGreen}{\textbf{1. For Known:}}\\ 
The Reversal Prompt and Benchmark Prompt should complement each other. 
Analyze the content and structure of both prompts to identify their accuracy, similarities and differences. 
Synthesize the key points and integrate them into a unified and coherent output. \\
\textcolor{Plum}{\textbf{2. For Unknown:}}\\ 
Extract a cognitive preference template $T$ from any inaccuracies in the LLM-taste prompt. Integrate meta-cognitive elements from the original prompt $P$ into this template to enhance $T$.\\
\textbf{Input:} \\
LLM-Taste Prompt:[LLM-Taste Prompt] \\
Benchmark Prompt:[Benchmark Prompt] \\ \textbf{Output:} [Output]
}
\end{tcolorbox}
\noindent\begin{minipage}{\textwidth}
\captionof{figure}{Prompt for CPM (Known/Unknown)}
\vspace{-1em}\label{fig:cpm}
\end{minipage}
\end{center} 

\clearpage
\section{Case Study}
\label{sec:case}
\subsection{Case Study for Known Task (Game of 24)}
\subsubsection{$P_{final}$ Case study for \texttt{Game of 24}}
\label{sec:knowncase}
\begin{center}
\begin{tcolorbox}[width=\textwidth,colback=white, colframe=NavyBlue, title=$P_{final}$ Case study for known task after CPM]
{\slshape 
\textbf{Task Defination}\\
The task is to find a feasible mathematical expression using the four input numbers (a, b, c, d) in the order they are given, such that the result equals 24. The feasible solution should involve basic arithmetic operations (+, -, $\times$, /) and parentheses if necessary.\\
\textcolor{BlueGreen}{\textbf{Logical Pseudocode}}
\begin{enumerate}[label={}, left=0pt]
    \item For each permutation $P(A, B, C, D)$ in permutations of [A, B, C, D]:
    \item For each operation $op_n \in (+, -, \times, \div)$, where $n \in \{0, 1, 2, 3\}$:
    \begin{itemize}[left=0pt]
        \item For each permute $\exists(Expression == 24) \lor (Expression = permute[0] \; op_1 \; permute[1] \\ \; op_2 \; permute[2] \; op_3 \; permute[3])$:
        \item If $\exists \text{(Expression} == 24)$, then:
        \begin{itemize}[left=0pt]
            \item Print(Expression)
            \item Terminate the process.
        \end{itemize}
        \item Else, continue checking the next permutation.
    \end{itemize}
    \item If no solution is found, print "No feasible solution exists.".
\end{enumerate}

\textcolor{Plum}{\textbf{Case Examples:}}

\textbf{Example 1:}
Input: 4 6 7 1
Output: One feasible solution (e.g., $6/(1-(4/7)) = 24$).

\textbf{Example 2:}
Input: 4 2 2 1
Output: "No feasible solution exists."

\textcolor{Red}{\textbf{Input-Output Format:}} \\
\textbf{Input:} Four integers separated by space (e.g., "3 3 8 8"). \\
\textbf{Output:} If a feasible solution exists, output a string indicating one possible solution (e.g., "One feasible solution (e.g., $8 \div (3 - 8 \div 3) = 24$)."). If no solution exists, output the string "No feasible solution exists."

}

\end{tcolorbox}

\noindent\begin{minipage}{\textwidth}
\captionof{figure}{$P_{final}$ Case study for \texttt{Game of 24}}
\vspace{-1em}\label{fig:case1}
\end{minipage}

\end{center}

\clearpage
\subsubsection{Instantiation Case Study for \texttt{Game of 24} Using $P_{final}$}
\begin{center}
\begin{tcolorbox}[width=\textwidth,colback=white, colframe=NavyBlue, title=Instantiation Case Study for \texttt{Game of 24} Using $P_{final}$]
{\slshape 
\textcolor{NavyBlue}{\textbf{Input:}} 4 9 10 13 \\
\hrule 
\vspace{0.8em} 
\textcolor{BlueGreen}{\textbf{Steps:}} 
Let me compute possible expressions for this input that result in 24 using provided logic pseudocode.\\
1. Generate all permutations of the numbers (4,9,10,13): This gives 24 possible arrangements. \\
2. Apply operations: For each permutation, test all combinations of the operations (+, -, *, /) between the numbers. \\
3. Check if the result equals 24: Evaluate the expression and check if it equals 24. \\
4. Output the result: If a valid expression is found, print it. If not, output that no feasible solution exists. 
\begin{verbatim}
import itertools
inputs = [4,9,10,13]
operations = ['+', '-', '*', '/']
def evaluate_expression(expr):
    try:
        return eval(expr)
    except ZeroDivisionError:
        return None
permutations = itertools.permutations(inputs)
op_combinations = itertools.product(operations, repeat=3)
solutions = []
for perm in permutations:
    for ops in op_combinations:
        exprs = [
            f"({perm[0]} {ops[0]} {perm[1]}) {ops[1]} ({perm[2]} {ops[2]} {perm[3]})",
            f"({perm[0]} {ops[0]} ({perm[1]} {ops[1]} {perm[2]})) {ops[2]} {perm[3]}",
            f"{perm[0]} {ops[0]} ({perm[1]} {ops[1]} ({perm[2]} {ops[2]} {perm[3]}))",
            f"({perm[0]} {ops[0]} {perm[1]}) {ops[1]} ({perm[2]} {ops[2]} {perm[3]})",
            ...
        ]
        for expr in exprs:
            if evaluate_expression(expr) == 24:
                solutions.append(expr)
print(solutions)
\end{verbatim}```
}
\hrule 
\vspace{0.8em} 
\textbf{Output:} 4 * (9 + (10 - 13))=24
\end{tcolorbox}
\noindent\begin{minipage}{\textwidth}
\captionof{figure}{Instantiation Case Study for \texttt{Game of 24} Using $P_{final}$}
\vspace{-1em}\label{fig:case1_ins}
\end{minipage}
\end{center}

\clearpage
\subsection{Case Study for Unknown Task (MGSM)}
\subsubsection{$P_{final}$ Case study for \texttt{MGSM}}
\label{sec:unknowncase}
\begin{center}
\begin{tcolorbox}[width=\textwidth,colback=white, colframe=NavyBlue, title=$P_{final}$ Case study for unknown task after CPM]
{\slshape 
\textbf{Task Defination}\\
The task is to solve a math problem text with a description of a situation. Your task is to calculate the answer and provide it in the format specified.\\
\textcolor{BlueGreen}{\textbf{Logical Pseudocode}} \\
\begin{enumerate}[label={}, left=0pt]
    \item Input $u$ from the user in natural language.
    \item Parse $u$ for numbers and contextual information.
        \begin{enumerate}[label={}, left=0pt]
            \item Extract numbers $\alpha$, $\beta$, $\gamma$, etc.
            \item Identify contextual clues and operations, such as addition (+), subtraction (-), multiplication (*), division (/), and other implicit operations (e.g., percentages, halves, totals).
        \end{enumerate}
    \item For each identified operation:
        \begin{enumerate}[label={}, left=0pt]
            \item If operation is related to percentages, interpret $\times$ or division as needed.
            \item If operation is additive (+), perform $Add(\alpha, \beta)$.
            \item If operation is subtractive (-), perform $Subtract(\alpha, \beta)$.
            \item If operation is multiplicative (*), perform $Multiply(\alpha, \beta)$.
            \item If operation is divisive (/), perform $Divide(\alpha, \beta)$.
            \item If implicit operations (e.g., "half that much"), interpret accordingly.
        \end{enumerate}
    \item Handle complex structures, such as total amounts or remainders:
        \begin{enumerate}[label={}, left=0pt]
            \item Use context to evaluate remaining quantities (e.g., after consumption, sales).
            \item Apply operations in sequential order based on context.
        \end{enumerate}
    \item Calculate the final result based on all interpreted operations and numbers.
    \item Return the final result as a natural number or in currency (if applicable).
\end{enumerate}

\textcolor{Plum}{\textbf{Case Examples:}}

\textbf{Example 1:} \\
\textbf{Input:} A robe takes 2 bolts of blue fiber and half that much white fiber. How many bolts in total does it take?	 \\
\textbf{Output:} 3

\textcolor{Red}{\textbf{Input-Output Format:}} \\
\textbf{Input:} A natural language string describing a scenario that involves certain arithmetic operations. \\
\textbf{Output:} A single integer or float number that is the result of the operations described in the scenario.

}
\end{tcolorbox}
\noindent\begin{minipage}{\textwidth}
\captionof{figure}{$P_{final}$ Case study for \texttt{MGSM}}
\vspace{-1em}\label{fig:case2}
\end{minipage}
\end{center}

\clearpage
\subsubsection{Instantiation Case Study for \texttt{MGSM} Using $P_{final}$}
\begin{center}
\begin{tcolorbox}[width=\textwidth,colback=white, colframe=NavyBlue, title=Instantiation Case Study for \texttt{MGSM} Using $P_{final}$]
{\slshape 
\textcolor{NavyBlue}{\textbf{Input:}} Paul is at a train station and is waiting for his train. He isn’t sure how long he needs to wait, but he knows that the fourth train scheduled to arrive at the station is the one he needs to get on. The first train is scheduled to arrive in 10 minutes, and this train will stay in the station for 20 minutes. The second train is to arrive half an hour after the first train leaves the station, and this second train will stay in the station for a quarter of the amount of time that the first train stayed in the station. The third train is to arrive an hour after the second train leaves the station, and this third train is to leave the station immediately after it arrives. The fourth train will arrive 20 minutes after the third train leaves, and this is the train Paul will board. In total, how long, in minutes, will Paul wait for his train? \\
\hrule 
\vspace{0.8em} 
\textcolor{BlueGreen}{\textbf{Steps:}} \\
\textbf{Step 1: Symbol Information} \\
1. $\alpha$ (first train arrival) = 10 minutes \\
2. $\beta$ (first train stay) = 20 minutes \\
3. $\gamma$ (second train stay as a quarter of the first train's stay) = $\frac{\beta}{4}$ \\
4. $\delta$ (time between second train departure and third train arrival) = 60 minutes \\
5. $\varepsilon$ (fourth train arrival after third train leaves) = 20 minutes 

\textbf{Step 2: Identify Contextual Clues and Operations} \\
1. First train departure time = $\alpha + \beta$ \\
2. Second train arrival time = First train departure time + 30 minutes \\
3. Second train stay time = $\gamma$ \\
4. Third train arrival time = Second train departure time + $\delta$ \\
5. Fourth train arrival time = Third train departure time + $\varepsilon$ 

\textbf{Step 3: Perform Calculations Based on Identified Operations} \\
1. First train departure time = Add($\alpha,\beta) = 10 + 20 = 30$ minutes \\
2. Second train arrival time = Add(First train departure time,30) = $30 + 30 = 60$ minutes \\
3. Second train stay time = Divide(${\beta},4) = \frac{20}{4} = 5$ minutes \\
4. Second train departure time = Add(Second train arrival time ,Second train stay time) = $60 + 5 = 65$ minutes \\
5. Third train arrival time = Add(Second train departure time,$\delta$) = $65 + 60 = 125$ minutes \\
6. Third train departure time = Third train arrival time (since it leaves immediately) = 125 minutes \\
7. Fourth train arrival time = Add(Third train departure time, $\varepsilon$) = $125 + 20 = 145$ minutes 

\textbf{Step 4: Calculate the Final Result} \\
\text{Final Result: Fourth train arrival time = 145 minutes}
\\
}
\hrule 
\vspace{0.8em} 
\textbf{Output:} 145 minutes
\end{tcolorbox}
\noindent\begin{minipage}{\textwidth}
\captionof{figure}{Instantiation Case Study for \texttt{MGSM} Using $P_{final}$}
\vspace{-1em}\label{fig:case2_ins}
\end{minipage}
\end{center}

\clearpage
\section{Instructions for Human Evaluation}
\label{sec:instruction}

To evaluate the cognitive preference consistency  $Con_{cog}$ between LLMs-taste prompt for task cognitions across different tasks, three professional annotators adhere to the following guidelines:
\paragraph{\textbf{Criteria.}}\textbf{(1) Style Consistency}: How well responses maintain a consistent stylistic tone for thinking across different tasks. \textbf{(2) Coherence}: Internal logic and smooth flow of the response.

\paragraph{\textbf{Scoring.}} Responses are evaluated on a scale from 0 to 5. A score of \textbf{4-5} signifies exemplary style consistency, while \textbf{3-4} indicates strong consistency with minor variations. Scores of \textbf{2-3} represent adequate style but with noticeable inconsistencies, and \textbf{1-2} denote inconsistent style across tasks. A score of \textbf{0-1} reflects complete inconsistency. Prior to the formal evaluation, several pre‑scored sample responses covering every score band will be provided. These examples serve to calibrate annotators’ judgments and to align their understanding of the rubric.

\end{document}